\documentclass[letterpaper, 10 pt, conference]{ieeeconf}  

\IEEEoverridecommandlockouts                              

\usepackage{color}
\usepackage{epsfig}
\usepackage{graphicx}
\usepackage{algorithm,algorithmic}

\usepackage{adjustbox}
\usepackage{array}
\usepackage{booktabs}
\usepackage{colortbl}
\usepackage{float,wrapfig}
\usepackage{framed}
\usepackage{hhline}
\usepackage{multirow}
\usepackage{makecell}
\usepackage[percent]{overpic}

\usepackage{amsmath,amsfonts,amssymb}
\usepackage{amsthm} 
\usepackage{bm}
\usepackage{nicefrac}
\usepackage{microtype}
\usepackage{contour}
\usepackage{courier}

\usepackage{changepage}
\usepackage{extramarks}
\usepackage{fancyhdr}
\usepackage{lastpage}
\usepackage{setspace}
\usepackage{soul}
\usepackage{xspace}
\usepackage{cuted}
\usepackage{fancybox}
\usepackage{afterpage}
\usepackage{gensymb}

\usepackage[breaklinks=true,colorlinks,backref=True]{hyperref}
\hypersetup{colorlinks,linkcolor={black},citecolor={MSBlue},urlcolor={magenta}}
\usepackage{url}
\usepackage{quoting}
\usepackage{epigraph}

\usepackage{enumerate}
\usepackage{paralist,tabularx}
\usepackage{comment}
\usepackage{pdfpages}
\usepackage{caption}  

\usepackage{pifont}

\usepackage{MnSymbol}

\usepackage[para]{footmisc}
\usepackage{cite}

\makeatletter
\DeclareRobustCommand\onedot{\futurelet\@let@token\@onedot}
\def\@onedot{\ifx\@let@token.\else.\null\fi\xspace}

\def\etal{et al\onedot}

\makeatother

\definecolor{MyDarkBlue}{rgb}{0,0.08,1}
\definecolor{MyDarkGreen}{rgb}{0.02,0.6,0.02}
\definecolor{MyDarkRed}{rgb}{0.8,0.02,0.02}
\definecolor{MyDarkOrange}{rgb}{0.40,0.2,0.02}
\definecolor{MyPurple}{RGB}{111,0,255}
\definecolor{MyRed}{rgb}{1.0,0.0,0.0}
\definecolor{MyGold}{rgb}{0.75,0.6,0.12}
\definecolor{MyDarkgray}{rgb}{0.66, 0.66, 0.66}
\definecolor{MyPink}{rgb}{1, 0.75, 0.79}
\definecolor{GreenStarColor}{rgb}{0.54, 0.84, 0.41}
\definecolor{MSBlue}{rgb}{0, 0.35, 0.49}

\def\OURS{VibeCheck\xspace}

\title{\huge \bf \OURS: Using Active Acoustic Tactile Sensing\\for Contact-Rich Manipulation}

\author{
Kaidi Zhang\authorrefmark{1}\authorrefmark{2}, %
Do-Gon Kim\authorrefmark{1}\authorrefmark{2}, %
Eric T. Chang\authorrefmark{1}\authorrefmark{2},
Hua-Hsuan Liang\authorrefmark{3}, %
Zhanpeng He\authorrefmark{3},\\%
Kathryn Lampo\authorrefmark{2},
Philippe Wu\authorrefmark{2},
Ioannis Kymissis\authorrefmark{4}, %
Matei Ciocarlie\authorrefmark{2} \\ %
\thanks{}  
}

\begin{document}

\twocolumn[{%
\renewcommand\twocolumn[1][]{#1}%
\maketitle
\begin{center}
    \centering
    \vspace{-3em}
    \captionsetup{type=figure, font=small}
    \includegraphics[width=\textwidth]{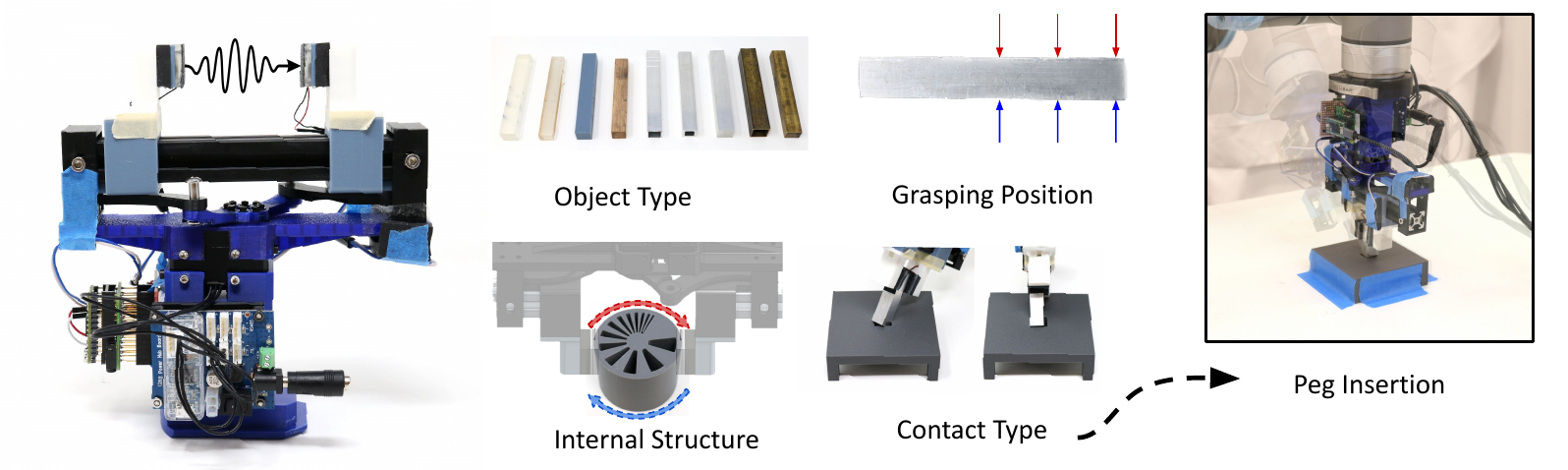}
    \captionof{figure}{\textbf{Overview of \OURS}. Two piezoelectric fingers on a parallel gripper send and receive acoustic signals through an object, providing information about resonant frequencies and object state (``active acoustic sensing''). By using this information to estimate the type of contact a rod is making with the environment, we demonstrate a peg insertion task using \textit{only} acoustic tactile feedback.}
    \label{fig:candy}
\end{center}
}]
{
  \renewcommand{\thefootnote}%
    {\fnsymbol{footnote}}
  \footnotetext[1]{equal contributions and joint-first authorship.}
  \footnotetext[2]{Dept. of Mechanical Engineering,} %
  \footnotetext[3]{Dept. of Computer Science,} %
  \footnotetext[4]{Dept. of Electrical Engineering, Columbia University, New York, NY 10027, USA.} %
  \footnotetext[0]{This work was supported by a NASA Space Technology Graduate Research Opportunity.}
  \footnotetext[0]{Correspondence email: {\tt\small eric.chang@columbia.edu}}
}
\thispagestyle{empty}
\pagestyle{empty}

\begin{abstract}
The acoustic response of an object can reveal a lot about its global state, for example its material properties or the extrinsic contacts it is making with the world. In this work, we build an active acoustic sensing gripper equipped with two piezoelectric fingers: one for generating signals, the other for receiving them. By sending an acoustic vibration from one finger to the other \textit{through} an object, we gain insight into an object's acoustic properties and contact state. We use this system to classify objects, estimate grasping position, estimate poses of internal structures, and classify the types of extrinsic contacts an object is making with the environment. Using our contact type classification model, we tackle a standard long-horizon manipulation problem: peg insertion. We use a simple simulated transition model based on the performance of our sensor to train an imitation learning policy that is robust to imperfect predictions from the classifier. We finally demonstrate the policy on a UR5 robot with active acoustic sensing as the \textit{only} feedback. Videos can be found at \href{https://roamlab.github.io/vibecheck}{https://roamlab.github.io/vibecheck}.

\end{abstract}
\section{Introduction}
\label{sec:intro}

Tactile sensors typically acquire data by making and breaking contact with the environment, providing local information about contact points and surfaces.
While many tasks benefit from knowing where contact occurs and understanding local geometry and forces, passive, intrinsic contacts (i.e. contacts between a sensor and the environment) are limited by the sensing area of the sensor and leave a lot of useful information unutilized. 
For instance, material properties and extrinsic contacts (i.e., contacts an object makes with the environment, not the sensor) are difficult to infer.

One way to provide such information to a local tactile sensor is to introduce an active component \cite{kim_active_2022, liu_sonicsense_2024}, conceptually similar to how a child would shake or tap a boxed present to figure out its contents. In other words, one can introduce a signal into the system and observe the response.

This is the principle of active acoustic sensing, in which an active emitter element (i.e. a speaker) emits an acoustic signal and a receiver element (i.e. a microphone) records the response \cite{zoller_active_2020, mason_acoustic_2024, kuboaudiotouch2019, funato_estimating_2017, onotouch2013, zhangefingerping2018, sandykbayeva_vibrotouch_2022, wu_vibmilk_2024}. In the context of a multi-fingered robotic gripper, this means sending an acoustic signal through an object from one finger to another. 
This signal can be useful for learning about object composition and geometry, extrinsic contact state (i.e. how the object is making contact with the environment), and even internal structure or mass, all qualities that typical tactile and vision modalities struggle with \cite{lu_active_2023, yi_visual-auditory_2024}.

In this paper, we build a parallel gripper containing a piezoelectric speaker and contact microphone. We use the same hardware for the two fingers, taking advantage of the piezoelectric property that allows the transducer to be both an actuator and a sensor. 
We use this system to explore the capabilities of active acoustic sensing for object classification, grasping position estimation, pose estimation from internal structure, and contact state classification.
Moving beyond static classification (i.e., isolated classification tasks), our key result is using this sensing modality to tackle a long-horizon manipulation task---the classic problem of peg insertion---learning a policy that successfully inserts a peg using active acoustic sensing as the only sensing modality. The primary contributions of this work are as follows:
\begin{itemize}
\item To our knowledge, we are the first to go beyond static classification and demonstrate that a long-horizon manipulation task, in this case a peg insertion task, can be learned using active acoustic sensing as the \textit{only sensing modality}.
\item To accomplish this, we build on previous work and recreate a hardware platform for active acoustic sensing between two fingers \cite{lu_active_2023}. We streamline the design by using the same component type as both emitter and receiver.
\item We demonstrate strong performance on tasks shown in previous work, including object classification, grasping point estimation, and contact state classification \cite{lu_active_2023, yi_visual-auditory_2024}. We also demonstrate an additional active acoustic sensing capability: estimating pose from internal structure. 
\end{itemize}

\section{Related Work}
\label{sec:related}

A few works have demonstrated how active acoustic sensing can be used in robotics. 
In soft pneumatic actuators (SPAs), a speaker and receiver system can be placed within the air cavity of the actuator, measuring the change in the acoustics due to elastomer deformation. 
This type of system can be an effective way to measure contact location and actuator shape \cite{zoller_active_2020, chandrasiri_transferable_2022, takaki_acoustic_2019, mikogai_contact_2020, randika_estimating_2021}. Beyond SPAs, a similar system can be used to create a low-cost tactile skin. 
This work places two piezoelectric contact microphones on a robot arm to do proximity sensing \cite{fanaurasense2021} and contact localization \cite{fanenabling2022} over the entire arm, effectively turning the arm into a large tactile skin. Active acoustic sensing can also be implemented with a single transducer. 
SonicFinger measures proximity using a single piezoelectric transducer to emit a signal into the environment and measure the returning signal \cite{rupavatharam_sonicfinger_2023}.

In manipulation, the more common use of acoustic sensors is passive contact microphones, which act as a dynamic tactile sensing modality. For example, SonicSense uses a multi-fingered hand equipped with contact microphones along with a heuristic-based exploration policy for container contents identification, 3D shape reconstruction, and object re-identification \cite{liu_sonicsense_2024}. Other works use contact microphones as tactile feedback for motor learning \cite{liu2024maniwav, mejia_hearing_2024, lee_sonicboom_2024, du_play_2022}. In comparison to our work, these works rely on dynamic interaction to create a signal and cannot measure object state without a stimulus from a motor or the environment.

A pair of studies have used active acoustic sensing in a parallel jaw gripper to measure object states, as we do in this work \cite{lu_active_2023, yi_visual-auditory_2024}. 
These studies demonstrate material classification, grasping position estimation, and contact state measurements in the presence of occlusion. 
They are the first to show active acoustic sensing can be useful in a gripper, where the fingers are ``communicating'' with each other through acoustics. 
However, Yi and Lee \etal only show how acoustic sensing, as a complementary sensing modality to visual sensors, can improve extrinsic contact identification performance.
In this study, we streamline the design by using the same component type as both emitter and receiver, and go beyond static classification to also demonstrate a complete control policy for a standard long-horizon manipulation task relying exclusively on active acoustic sensing. 

\section{Hardware Platform}
\label{sec:method}

\begin{figure}
\centerline{\includegraphics[width=0.95\linewidth]{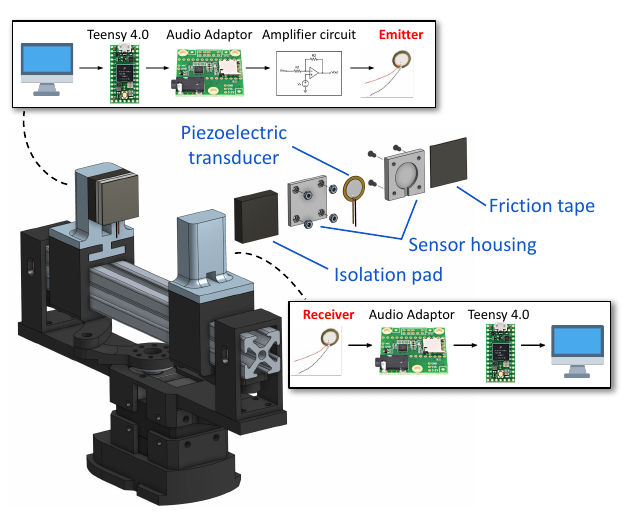}}
\caption{Design of our active acoustic tactile sensor on a parallel gripper. While the hardware for each finger is the same, one acts as a speaker and the other as a receiver.}
\label{fig:hardware}
\end{figure}

\subsection{Sensing Principle}
In our parallel gripper active acoustic sensing system, one finger contains an emitter element (i.e., a speaker) and the other finger contains a receiver element (i.e., a contact microphone). When an object is held between the two fingers, the speaker emits a known acoustic waveform, transmitting vibration to the object which is sensed by the receiving element. 
As a result, the received waveform is highly dependent on the acoustic resonance properties of the object and can capture information about material composition, shape, mass distribution, and extrinsic contacts.

\subsection{Hardware Design}
Building off of previous work, our hardware platform is inspired by Lu and Culbertson \cite{lu_active_2023} with a few modifications that enhance the robustness and consistency of the sensor. These design decisions were based on empirical analysis of both the raw signals produced and performance in downstream tasks.

\textit{1) Piezoelectric Transducer:} On each of the fingers, we embed an Adafruit piezoelectric disk (Product ID: 1740, resonant frequency: 8 kHz). We take advantage of the fact that piezoelectrics can both convert strain into voltage as well as voltage into strain and use the same type of transducer as both the speaker and receiver. 
We find that using this piezoelectric disk as the speaker performed equally well or better than using a designated bone conduction actuator. Unlike conventional audio microphones, the piezoelectric receiver acts as a contact microphone and is relatively immune to ambient noise. To enhance compactness and sensitivity, we remove the sensor's plastic housing. The unenclosed sensor has a 0.2 mm thickness and 6 mm radius, allowing the assembled finger to be thin and compact.

\textit{2) Sensor Housing:} The transducers are mounted on the fingers with 3D-printed housings (Formlabs Clear V4 resin) and sorbothane isolation pads to reduce parasitic vibrations (Fig. \ref{fig:hardware}). 
We find these isolation pads important in absorbing vibrations and noise generated by the robot’s movement and transmitted through the gripper body. 
Lastly, we add replaceable high-density polyethylene strips (“Non-Slip Grip Strips” by CatTongue) to the fingertips to increase friction at the contact surface.

\begin{figure}[t]
\centerline{\includegraphics[width=0.95\linewidth]{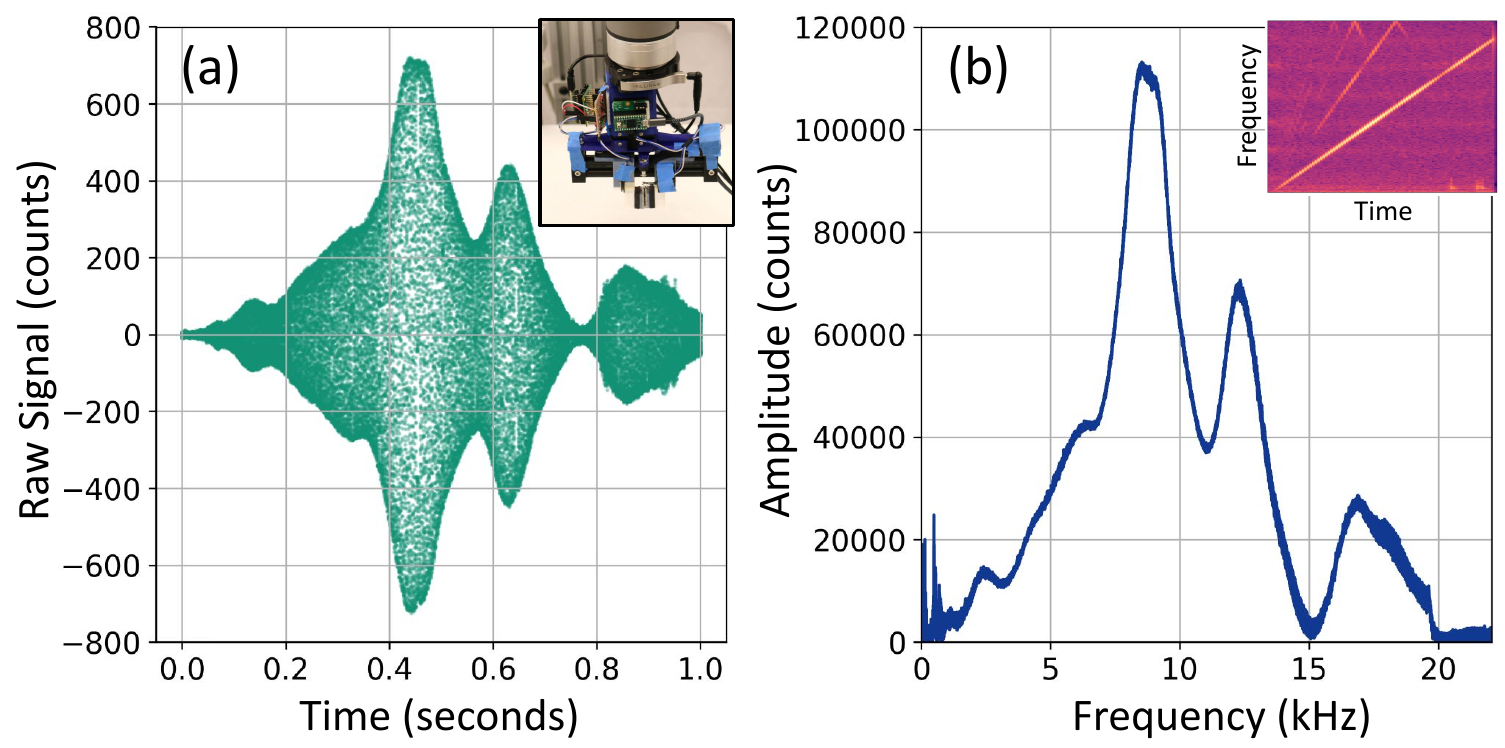}}
\caption{Time domain (\textit{a}) and frequency domain (\textit{b}) signals when the gripper is fully closed (\textit{a} inset), as well as a spectrogram of the signal over 1 second (\textit{b} inset). The major peak around 8 kHz is consistent with the piezoelectric disk's reported resonant frequency.}
\label{fig:rawsignal}
\end{figure}

\textit{3) Electronics:} Two Teensy 4.0 microcontrollers with audio adapter boards control the signal generation, sampling, and communication with a computer. The generated signal is passed through an inverting amplifier circuit (gain = 1.5) and is biased such that the resulting signal is centered at 2.5 V and has an amplitude of 2.4 V. The received signal is unamplified and sampled at 44.1 kHz. The Teensys communicate with the computer using micro-ROS \cite{belsare2023microrOS}.

\subsection{Sensing Procedure and Data Preprocessing}

The sensing process involves generating an excitation on one finger while simultaneously recording the signal received by the other finger. During each sensing cycle, a sinusoidal signal is linearly swept from 20 Hz to 20 kHz for 1 second. Linear sweep signals increase in frequency at a constant rate over time, enabling us to focus on a range of frequencies during offline analysis. Before training, we retain only the first 42,000 points of the sweep (due to data corruption issues), effectively terminating the sweep at 19.029 kHz.

We convert the raw time domain data to the frequency domain using fast Fourier transform (FFT). We then use kernel principal component analysis (kernel PCA) with a cosine kernel to reduce the dimensionality of the data before model training. 
We further elaborate on the selection of principle components in the following section (Sec. \ref{sec:experiments}).
An example of the data when the gripper is fully closed is shown in Fig. \ref{fig:rawsignal} in both the time and frequency domain.


\section{Estimating Object State and Properties with Acoustic Features}
\label{sec:experiments}

\begin{figure}[t]
\centerline{\includegraphics[width=0.75\linewidth]{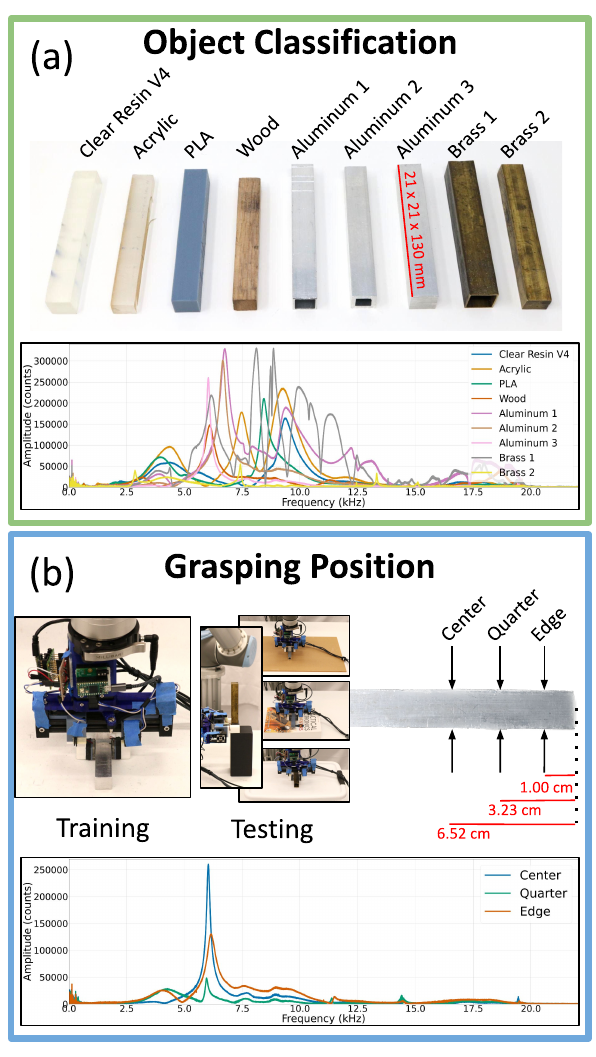}}
\caption{Setup for object type (\textit{a}) and grasping position (\textit{b}) experiments and examples of signals after FFT. All training data is collected with the rod resting on a table. We show generalization on test sets collected on new surfaces and object orientations.}
\label{fig:experiment_setup1}
\end{figure}

\begin{figure} [ht!]
\centerline{\includegraphics[width=0.75\linewidth]{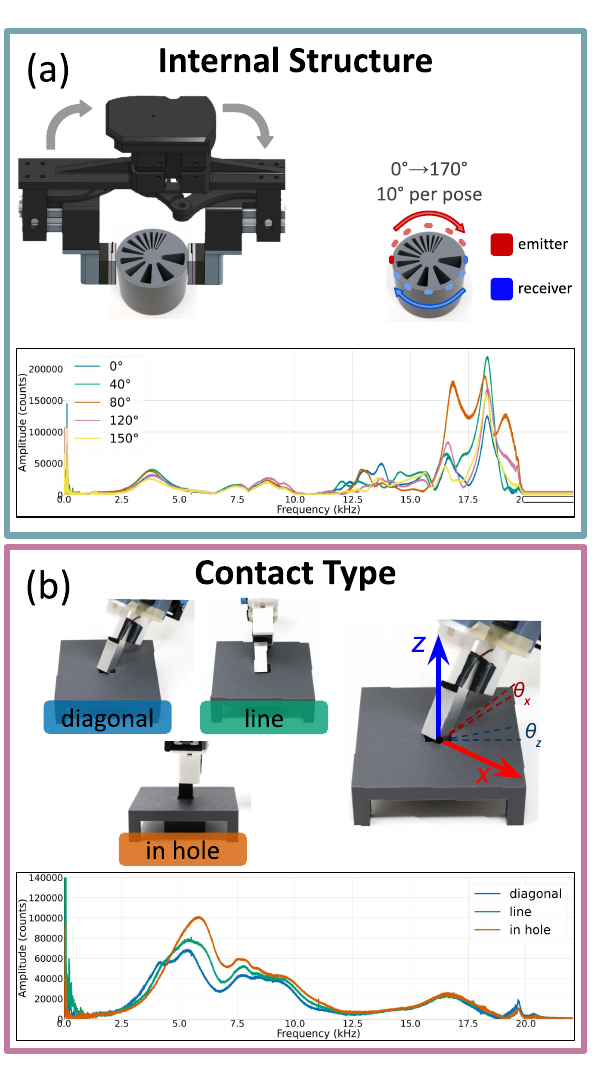}}
\caption{Setup for internal structure (\textit{a}) and contact type (\textit{b}) experiments, with examples of signals after FFT.}
\label{fig:experiment_setup2}
\end{figure}

We investigate the types of information that our sensor can capture by training a model for each of the tasks outlined in Fig. \ref{fig:experiment_setup1} and \ref{fig:experiment_setup2}. 
For all tasks, the dataset is divided into training, validation, and test sets. 
Although each subset may originate from multiple data collection sessions, we ensure that the test sets are derived from entirely separate sessions to account for distribution shifts between sessions, while the validation sets are sampled from the same datasets as the training set with a 9:1 split. 
For each task, we train a multi-layer perceptron (MLP) model of size (400, 250, 100) to do classification or regression.

\subsection{Object Classification}
We classify 9 rods of different materials and/or geometries (Fig. \ref{fig:experiment_setup1}a). We collected rods as close to $21 \times 21 \times 130$ mm as possible, however, we recognize that some of the rods differ slightly from these dimensions.  
We collect 100 and 25 training and test samples per object (900 and 225 total), respectively, grasping in the center of the object with the gripper attached to a UR5 robot arm and the object resting on a table. 
We collect 5 samples for each grasp of the object before opening the gripper and regrasping. 

As shown in Table \ref{tab:clf_results}, when testing in similar conditions to the training data, from a single grasp we show 100\% classification accuracy across 9 classes, demonstrating the sensor's ability to distinguish diverse materials and hollow/non-hollow geometries. 
To evaluate the robustness of our sensor and classifier, we also evaluate the object classifier under two unseen settings during training data collection by placing our objects on different surface materials and grasping the object in different orientations (\texttt{New Surface} and \texttt{New Orientation} in Table \ref{tab:clf_results}, 25 samples per condition/object).

Contacts with different surfaces were collected by placing the objects on an acrylic sheet, a paperback textbook, and a plastic container.
We find that the contact surface can change how vibrations are damped and can slightly shift the resonant frequencies. Contacts with different orientations were collected by holding the object in the air and rotating $+/- 45\degree$ about both horizontal plane axes. This removes any contact surface and changes how the weight of the object is distributed between the two fingers. We do not include \texttt{Brass 2} in this test set as our gripper was not strong enough to hold it in the air.
Our results show that the trained classifier can generalize to these conditions outside of the training distribution. 

In this task, we perform dimensionality reduction using kernel PCA with five principal components (PCs), as this selection yields the best performance for both in-distribution and unseen test cases. 
Notably, we achieve 100\% accuracy for in-distribution testing using only three PCs, which account for 75\% of the explained variance. However, to attain optimal performance on unseen test cases without degradation in in-distribution performance, 91\% of the explained variance is required.

\begin{table*}[h!]
\renewcommand{\arraystretch}{1.25}
\caption{Classification accuracy for object, grasping position, and contact type classification. ``In-distribution" is a test set that is collected in the same conditions as training data. ``New surface" and ``New orientation" are test sets in which the object is resting on a surface or in an orientation not seen in training. ``Interpolated poses" are test sets containing \texttt{diagonal} and \texttt{line} poses randomly sampled from within the range of training angles (``in-distribution") and \texttt{diagonal} poses randomly sampled from a wider range of angles than what is seen in training (``out-of-distribution")  (see Fig. \ref{fig:experiment_setup2}b and Section \ref{sec:experiments}).}
\label{tab:clf_results}
\centering
\begin{tabular}{c|c|c|c|c}
\multicolumn{2}{c|}{} & \multicolumn{3}{c}{\textbf{Accuracy}} \\ \hline
\multicolumn{2}{c|}{Frequency Range (kHz)}  & 0.02 - 22.05 & 0.02 - 9.19 & 9.19 - 22.05 \\ \hline \hline
\multirow{3}{*}{\textbf{Object}} & In-distribution  & 1.00 & 1.00 & 0.92  \\ \cline{2-5}
                               & New surface         & 1.00 & 1.00  & 0.87               \\ \cline{2-5}
                               & New orientation         & 1.00 & 1.00  & 0.63               \\ \hline
\multirow{3}{*}{\textbf{Grasping Position}} & In-distribution  & 1.00 & 1.00 & 0.97  \\ \cline{2-5}
                               & New surface        & 0.99 & 0.94  & 0.80               \\ \cline{2-5}
                               & New orientation         & 0.90 & 0.85  & 0.43               \\ \hline
\multirow{3}{*}{\textbf{Contact Type}}  & In-distribution  & 0.95 & 0.94 & 0.79  \\ \cline{2-5}
                               & Interpolated poses (in-distribution)        & 0.73 & 0.73  & 0.33               \\ \cline{2-5}
                               & Interpolated poses (out-of-distribution)         & 0.80 & 0.50  & 0.85               \\
\end{tabular}
\end{table*}

\subsection{Grasping Position Classification} Using the same set of materials, we classify 3 different grasping positions along each rod (Fig. \ref{fig:experiment_setup1}b). This type of task could be useful for reasoning about the relative pose of a known object under occlusion, particularly when searching for a stable grasp with equal weight distribution.

The 3 labels in this task are: \texttt{edge} (1.0 cm from the end), \texttt{quarter} (3.2 cm from the end, or a quarter of the way along the rod), and \texttt{center} (6.3 cm from the end, or approximately the lengthwise middle of the rod). We collect 100 and 25 training and test samples per point (2700 and 675 total), respectively, with the object and gripper positioned the same way as for the material type task and regrasping every 5 samples.

As shown in Table \ref{tab:clf_results}, when testing using the same distribution as training, we see 100\% accuracy across 3 classes. 
As in object classification, we also test on unseen surfaces and orientations. 
The process for new surfaces is the same, however, new orientations are not possible in the same way because the gripper is not strong enough to hold objects unsupported when grasping at the edge of the rods. 
Instead, we collect test samples by placing the objects upright on a plastic stand (Fig. \ref{fig:experiment_setup1}b).
Our results show that the model generalizes well to our out-of-distribution conditions.

Similar to object classification, we find that selecting the number of principal components that preserve 90\% of the explained variance yields the best performance across all test cases. While using more than 10 PCs can achieve 100\% accuracy on the in-distribution test set, restricting the model to 10 PCs helps prevent overfitting to the training dataset and ensures optimal performance on all unseen test cases.

\subsection{Pose Estimation from Internal Structure} 

\begin{figure} [!ht]
\centerline{\includegraphics[width=0.85\linewidth]{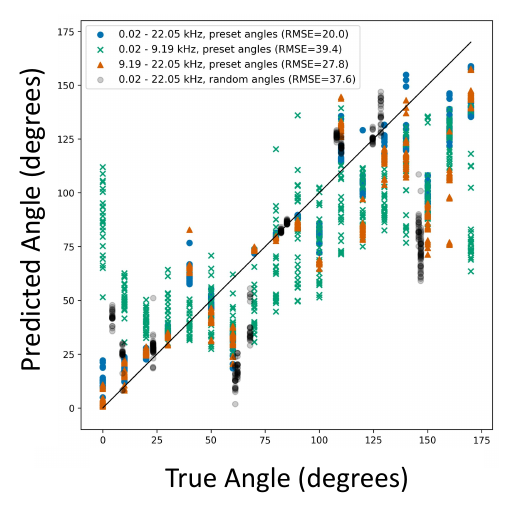}}
\caption{Regression results on the pose estimation from internal structure task using 3 frequency ranges. The grey circles denote randomly sampled poses that are \textit{not} seen in training.}
\label{fig:internal_results}
\end{figure}
We then investigate using our sensor to obtain information about the internals of an object that may not be visible from the exterior. 
Given a known object, we show that our sensor can reveal information about the relative pose of the object using only local contact information.
To explore this, we 3D-printed a polylactic acid (PLA) cylinder (Fig. \ref{fig:experiment_setup2}a). The cylinder is smooth around the outside but has an asymmetric internal geometry. The object is printed with the cylinder axis facing upwards and with 100\% infill. 

We investigate pose estimation of the cylinder about the vertical axis. We collect data at 10$\degree$ discretized poses around the cylinders. We only collect data from 0 to 170 degrees (18 poses) as the fingers repeat poses (with speaker and receiver flipped) beyond this. We collect 100 and 25 training and test samples per angle (1800 and 450 total), respectively, and regrasp the object every 5 samples. 

As shown in Fig. \ref{fig:internal_results}, we find that our acoustic sensor can capture some information about object pose relative to the gripper. Although a root mean squared error (RMSE) of 20.0$\degree$ is fairly large, we find that the model has a smaller error for most angles, with performance dropping at low and high angles.
Additionally, we test the model's performance on 16 randomly sampled angles between 0 and 170$\degree$ to test the model's performance on interpolated, unseen poses (Fig. \ref{fig:internal_results}). We are encouraged to find that the error for interpolated poses is of a similar scale to that for poses seen during training.
In this task, we find that using five principal components yields the best performance on the in-distribution test set, accounting for approximately 87\% of the explained variance.

\subsection{Contact Type Classification} 
For our final task, we classify three different types of contacts that an aluminum rod makes with the environment. 
We later use this classifier for a peg-insertion task, so the 3 types of contacts defined here are the types of contacts that might be experienced while inserting a rod into a square hole. 
As shown in Fig. \ref{fig:experiment_setup2}b, the three contacts are: \texttt{diagonal} ($0 < \theta_z < 90\degree$), \texttt{line} ($\theta_z=90, ~ \theta_x<90$), and \texttt{in-hole} ($\theta_z=90, ~ \theta_x=90$), when the rod is not making any contacts with the hole or the table. 
We use a $13.1 \times 12.8 \times 53.9$ mm rod for this task due to grasping force constraints during the insertion task. 
Meanwhile, the hole is 3mm expanded in each horizontal direction, which yields a 6$\degree$ error tolerance for the peg-insertion task.

We collect 1000 training samples and 200 test samples for each of \texttt{diagonal}, \texttt{line}, and \texttt{in-hole} contact types. As shown in Fig. \ref{fig:experiment_setup2}b, we limit data collection to ($45\degree \leq \theta_z \leq 90\degree, ~ \theta_x = 45\degree$) for \texttt{diagonal} states, and ($\theta_z = 90\degree, ~ 45\degree \leq \theta_x < 90\degree$) for \texttt{line} states, and collect data in $4.5\degree$ discretizations. We collect data by starting at $\theta_z = 45\degree$ and completing a handcrafted trajectory to the \texttt{in-hole} state. The trajectory consists of rotations about $z$ to the \texttt{line} state, followed by discrete rotations about $x$ to the \texttt{in-hole} state.

For each discrete pose, we begin with the gripper positioned a few centimeters above the platform, lower it until contact is made, and then raise the robot before determining the rotation action to the next pose. This approach is used to avoid large contact forces during rotation that could cause the rod to become dislodged from the gripper.

This method raises the question of how to ensure that the rod contacts the hole before a sensor reading is taken, as hard-coded $z$-axis positions are not robust to small variations in the grasp pose.
To address this, we analyze the receiver signal as the robot descends onto the platform and apply a threshold to determine the moment of contact.
We find this heuristic to be highly effective, and discard the occasional premature contact detection during data collection.
We also set a second threshold on the \textit{z} axis: if the robot moves too far downwards, we stop the robot and take a sensor reading. This is to catch the final state, when the robot will make no contact if the peg is properly aligned, as well as to act as a safety stop.

As shown in Table \ref{tab:clf_results}, our results show that our sensor can estimate the contact state of a rod with the environment, distinguishing between different contact states seen during a peg insertion task. 
We find that classification performance tends to be the worst in states that are on the border between \texttt{diagonal} and \texttt{line}, with a less significant drop at the border between \texttt{line} and \texttt{in hole}.

We also test this classifier in two scenarios: interpolated poses and out-of-distribution poses (Table \ref{tab:clf_results}). For the interpolated poses, we randomly sample 20 positions from each of \texttt{diagonal} and \texttt{line} states within the range of the training data (i.e. not discrete training poses). For out of distribution poses, we randomly sample 20 positions from the range ($40.5 \leq \theta_x \leq 171\degree,~ 9\degree \leq \theta_z < 90\degree$) (\texttt{diagonal} state). This set includes only \texttt{diagonal} states.

In this task, we observe that reducing the dimensionality of the training data to a level comparable to other tasks is challenging.
Our results show that using 500 PCs yields the best performance on the test set. 
Furthermore, we find that the contact type classifier is sufficiently robust for an insertion task, as discussed in the following section.

\section{Task learning}
\label{sec:task_learning}

\begin{figure*}[!t]
\centerline{\includegraphics[width=0.9\textwidth]{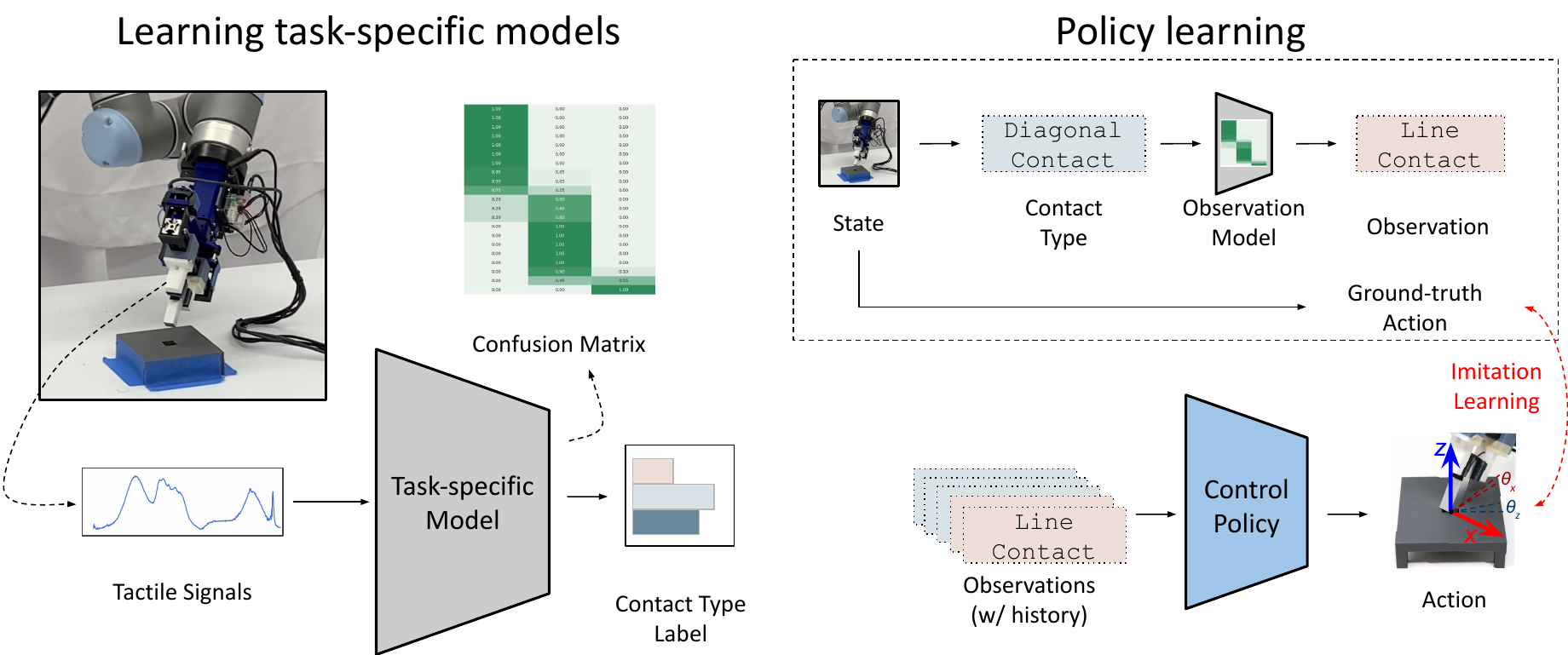}}
\caption{Learning task-specific classifiers and policy derivation. We use the probability distribution of the contact type classifier in simulation to train an imitation learning policy for peg insertion that incorporates history and is robust to imperfect contact state observations. The example confusion matrix shown here displays classifier test performance at each discrete data collection pose vs 3 contact labels.}
\label{fig:task_learning}
\end{figure*}

\subsection{Task Overview}
Using the contact type classifier, we demonstrate the capability of this sensor to be used as a stand-alone sensing modality in a long-horizon task: peg insertion. In this task, we aim to achieve a policy that can maneuver the rod into the hole starting from a random initial pose and using feedback solely from the contact type classifier. We use a Universal Robots UR5 robot equipped with our gripper for these experiments.

A simple approach is to design a handcrafted policy using heuristics that takes the classifier output as observations. Although our classifier performance is high overall, we empirically find that handcrafting policies is challenging due to imperfect predictions. A common strategy in robot learning to mitigate this issue is to incorporate an observation history. However, manually enumerating different possibilities in the observation history is impractical.

\subsection{Simulator and Training}
We instead use a simulated environment to train a policy that is robust to uncertain observations, which we then transfer directly to the robot. As discussed in Section \ref{sec:experiments}, we handcraft an ``ideal" trajectory consisting of incremental rotations about the $z$ axis, followed by incremental rotations about the $x$ axis until the rod is aligned with the hole. As a result, in the simulator, we have easy access to ground truth actions, since the ground truth contact type is always known.

At a high level, the simulator keeps track of 112 discrete robot poses (see poses used in data collection in Section \ref{sec:experiments}), where the action space contains 2 actions: move $+4.5\degree$ about the $x$ or $z$ axis. 
The observation contains a history of classifier outputs, which are sampled from a probability distribution constructed using the trained classifier. 
In the simulator, we use sampled observations and ground truth action pairs as demonstrations to train an imitation learning policy (Fig. \ref{fig:task_learning}).

More formally, given a classifier $h$, observed contact type (sampled from the test set confusion matrix of $h$) $c_{obs}$, and ground truth contact type $c_{gt}$, we construct a categorical distribution $p(c_{obs} | c_{gt})$ over the predicted labels.
To collect data for training, we use the handcrafted policy that uses ground-truth labels $c_{gt}$ as inputs and outputs $a_{expert}$.
Finally, we collect $\{c_{obs}, a_{expert}\}$ pairs for imitation learning, whose final product is a policy $\pi(a | c^0_{obs}, c^1_{obs}, ... c^n_{obs})$ that operates on $n=10$ steps history of observations.
To train the policy, we minimize the negative log-likelihood of the expert actions:
\[
\mathcal{L} = - \log \pi(a_{expert} | c^0_{obs}, c^1_{obs}, ... , c^n_{obs})
\]
In this work, $\pi$ is a categorical distribution since we use a discrete action space. 
However, our method can be extended to continuous action spaces by using continuous distributions. 
We consider a rollout successful if it reaches the insertion goal state within 50 steps.

\subsection{Rollout on UR5}
The final system designed for the peg-insertion task consists of three components: a heuristic-based collision detection mechanism, an external contact-type classifier, and a control policy that incorporates historical observations.
When executing a policy on the UR5, we use the same contact sensing and rotation procedure as during data collection, described in Section \ref{sec:experiments}.
To handle the occasional false positive contact detection, we treat an almost immediate contact detection as a false positive and repeat the detection process, while ignoring any others.

\subsection{Results}
The policy in simulation completes the insertion with $95\%$ success rate with uncertain observations. On the robot, we evaluate the policy in a few different settings:

\textit{1) Fixed Start Pose, $\theta_x = \theta_z = 45\degree$:} We first rollout the policy with a fixed start pose in the same conditions as data collection. Because the action space contains fixed angle increments, all visited discrete poses are in training if the policy takes perfect actions (i.e. rotates first in $z$, then in $x$). 

\textit{2) Interpolated Start Poses, $\theta_x = 45\degree, ~ 45\degree \leq \theta_z \leq 90\degree$:} Rather than starting at the discrete training pose, we also rollout the policy using start poses that are randomly sampled within the training data range. The 4.5$\degree$ action discretization is small enough to achieve final poses at which our model determines to insert the peg. 

\textit{3) Out-of-Distribution Start Poses, $40.5\degree \leq \theta_x \leq 81\degree, ~ 9\degree \leq \theta_z \leq 90\degree$:} Finally, we sample random start poses from a range outside of the training distribution. This includes all possible $\theta_z$ values up to the previous \texttt{line} state, and all possible $\theta_x$ values within the limitations of the gripper hardware (i.e. where it collides with the platform). 

Our results are shown in Table~\ref{tab:insertion_results}, and recordings of rollouts can be found in our \href{https://roamlab.github.io/vibecheck}{supplementary video}. For in-distribution cases, the success rate is particularly high for test case \textit{2)}. Even for poses that are significantly out-of-distribution, we still observe successful insertion 60\% of the time. We find this to be an encouraging result, as it suggests generalization of the policy and classifier to nearby poses.

\begin{table}[t!]
\renewcommand{\arraystretch}{1.3}
\caption{Success rate over 10 trials of insertion. $\theta_x = \theta_z = 45\degree$ indicates the same start conditions as training data collection. "Random Start Poses" indicates a randomly sampled start pose. We test the latter in 2 cases: when $\theta_x=45\degree$, keeping the insertion in the training distribution, and $40.5\degree \leq \theta_x \leq 81\degree$, where $\theta_x$ values are out of the training distribution (Fig. \ref{fig:experiment_setup2}).}
\label{tab:insertion_results}
\centering
\begin{tabular}{c|c|c}
    & \begin{tabular}[c]{@{}c@{}}$\theta_z=45\degree$\\Start Pose\end{tabular} & \begin{tabular}[c]{@{}c@{}}Random\\Start Poses\end{tabular} \\\hline
$\theta_x$ = 45\degree & 6/10  & 9/10   \\ \hline
40.5\degree  $\leq$ $\theta_x$ $\leq$ 81\degree & N/A & 6/10
\end{tabular}
\vspace{-1.5em}
\end{table}

\section{Discussions}
\label{sec:discussions}

\subsection{Robustness to external noises and vibrations}
A key limitation of acoustic-based sensors is their susceptibility to inconsistency in noisy environments. In this work, our design mitigates this issue by isolating the sensors from the robot using isolation pads, which reduce noise caused by vibrations from the robot arm. Furthermore, we test the robustness of our sensor signals to external noise by evaluating our models on an additional test set (300 samples) collected while a speaker added distractor signals (music measured at 75 dB ambient level) during the contact type classification task. Our results show that our contact type model can still achieve 87\% accuracy in the presence of this noise.

\subsection{Effective frequency ranges}
Frequency range selection is important for downstream tasks using active acoustic sensing in that it determines how fast we can complete sensing at a time step.
More crucially, for some of our tasks, the frequency range affects the performance of our models. 
For most test sets, while the full range yields better results, we find that $0.02$ kHz to $9.19$ kHz generally has less degraded performance, which is consistent with this range qualitatively containing more prominent frequency features. These findings suggest to us that if tuning to a specific task, a shorter sweep is often sufficient.

\subsection{Where does acoustic sensing fail?}
Despite the strong performance demonstrated in this work, we acknowledge several limitations of this sensing paradigm.
First, the signal response is influenced by many factors, some of which are informative while others introduce variability. 
For instance, the response can be affected by minor hardware adjustments or motor heating. 
Additionally, as our experiments illustrate, the acoustic response depends on geometry, material properties, and contact states, among other factors. However, these characteristics also make acoustic sensing a valuable complementary modality, providing rich information for fine-grained manipulation tasks.

In this vein, our learned models are specific to the object sets used in this work. While they exhibit some degree of generalization, as shown in Table \ref{tab:clf_results}, they do not generalize broadly to entirely unseen objects, nor is this expected. The signal response is influenced by the acoustic properties of the entire system as well as environmental disturbances. Consequently, achieving truly general acoustic sensing requires additional considerations when curating a diverse dataset.
However, in many practical applications, this limitation is not a concern, as reasonable assumptions about the environment can be made. For instance, acoustic sensing is well-suited for tasks such as sorting known objects or achieving stable grasps of a familiar object under occlusion, such as when reaching into a bag.

\subsection{Integration with other tactile sensors}
Active acoustic sensing provides unique features of tactile feedback, however in a more general robotic system one would likely have additional design goals and constraints. 
We do think that active acoustic sensing and other tactile modalities can coexist.  
One possible practical implementation is to have a single high-frequency actuator paired with other fingers containing innate high-frequency dynamic tactile sensing modalities acting as the receivers. 
Other types of tactile sensors, grasp stability, and manipulation skills in general benefit from the compliance of the finger -- since adding finger surface compliance (e.g. an elastomer) attenuates vibrations, one would likely need a more powerful actuator to coincide with these designs.

\section{Conclusion}
\label{sec:conclusion}

In this work, we show that active acoustic sensing alone can be used as feedback for a long-horizon manipulation task: peg-insertion. We also explore the capabilities of active acoustic sensing for extracting information about object material and geometry, grasping pose, internal structure, and extrinsic contact type. Interestingly, we find that for many tasks, we can use kernel PCA to find and use just a few (5-10) principal components that explain about 90\% of the variance, helping the models generalize to different test conditions. We hope these results will be useful to those considering the benefits and drawbacks of active acoustic tactile sensing as a sensing modality for manipulation.
\section{Acknowledgements}
\label{sec:ack}

We thank Pedro Piacenza, Trey Smith, Brian Coltin, Peter Ballentine, and Ye Zhang for insightful discussion.

\bibliographystyle{IEEEtran}
\bibliography{references}

\addtolength{\textheight}{-12cm}   
\end{document}